%% file: main.tex
\documentclass[conference]{IEEEtran}
\IEEEoverridecommandlockouts
\usepackage{cite}
\usepackage{amsmath,amssymb,amsfonts}
\usepackage{algorithm}
\usepackage{algpseudocode}

\usepackage[spaces,hyphens]{xurl} 
\usepackage[colorlinks,allcolors=blue]{hyperref}

\usepackage{graphicx}
\usepackage{epstopdf}
\usepackage{textcomp}
\usepackage{xcolor}
\def\BibTeX{{\rm B\kern-.05em{\sc i\kern-.025em b}\kern-.08em
    T\kern-.1667em\lower.7ex\hbox{E}\kern-.125emX}}

\usepackage{fifo-stack}
\FSCreate{colors}{black}
\makeatletter
    \let\old@color\color
    \renewcommand\color[1]{\FSPush{colors}{#1}\old@color{#1}}
    \newcommand\colorend{\FSPop{colors}\old@color{\FSTop{colors}}}
\makeatother

\begin{document}

\title{Force-directed graph embedding with hops distance\\
}

\author{\IEEEauthorblockN{Hamidreza Lotfalizadeh}
\IEEEauthorblockA{\textit{School of Electrical and Computer Engineering} \\
\textit{Purdue University}\\
West Lafayette, IN, USA \\
HLOTFALI@Purdue.edu}
\and
\IEEEauthorblockN{Mohammad Al Hasan}
\IEEEauthorblockA{\textit{School of Informatics, Computing and Engineering} \\
\textit{Indiana University Indianapolis}\\
Indianapolis, IN, USA \\
ALHASAN@IUPUI.edu}
}

\maketitle

\begin{abstract}
Graph embedding has become an increasingly important technique for analyzing graph-structured data. By representing nodes in a graph as vectors in a low-dimensional space, graph embedding enables efficient graph processing and analysis tasks like node classification, link prediction, and visualization. In this paper, we propose a novel force-directed graph embedding method that utilizes the steady acceleration kinetic formula to embed nodes in a way that preserves graph topology and structural features. Our method simulates a set of customized attractive and repulsive forces between all node pairs with respect to their hop distance. These forces are then used in Newton's second law to obtain the acceleration of each node. The method is intuitive, parallelizable, and highly scalable. We evaluate our method on several graph analysis tasks and show that it achieves competitive performance compared to state-of-the-art unsupervised embedding techniques.
\end{abstract}


\title{Graph Embedding Techniques}
\author{Anonymous}

\date{\today}
\maketitle

\section{Introduction}

In graph theory, a graph is a mathematical structure that represents relationships between objects. Objects and relationships between pairs of objects are represented by a set of vertices and edges respectively. Another conventional name for vertex and edge is node and link which are used interchangeably in this paper. Graphs are the widely used data structures for representing relationships between entities in various domains of applications, such as social networks, biological networks, knowledge graphs, and communication networks. Analyzing graphs can provide valuable insights into these domains. Nevertheless, working with graph data also poses significant computational challenges due to its complex, interconnected structure. 

A key challenge when analyzing graphs is developing useful vector representations of the nodes, which can enable downstream machine learning tasks like node classification, link prediction, and clustering \cite{perozzi2014DeepWalk,cao2015grarep,grover2016Node2vec}. 

Graph embedding has emerged as an effective technique for converting graph-structured data into a form that is more amenable to analysis and computation. The key idea is to represent each node in the graph as a vector in a low-dimensional space. In other words, the objective of graph embedding on a graph with $n$ nodes is to place each node in a d-dimensional space, with $d \ll n$ such that the placements reflect important graph structure and connectivity information. The embedding is constructed to preserve node proximity, which refers to the proximity or similarity between a pair of nodes based on the graph structure. Nodes that are "close" in terms of hop distance in the original graph topology should be embedded closer together in the vector space.

Recent work has shown that low-dimensional embeddings of nodes in large graphs can encode useful information about network structure and node metadata \cite{tang2015line,wang2016structural,perozzi2014DeepWalk,cao2015grarep,grover2016Node2vec,hamilton2016diachronic}. The basic methodology behind these node embedding approaches is to use dimensionality reduction techniques tailored for graph structures to distill high-dimensional graph information into a dense vector representation for each node.

Graph embedding methods can be categorized into two major sets of supervised and unsupervised embedding methods. In unsupervised embedding, the only information available during the embedding process is the node connectivity information in terms of edges. In supervised embedding however, we also may have other information such as node and edge features.

In this work, we present a novel unsupervised graph embedding method based on force-directed objective functions. Force-directed approaches have been extensively used for graph drawing and visualization, where node positions are optimized to reflect graph topology. We adopt the force-directed paradigm to learn graph embeddings that preserve structural proximity between nodes. Our method simulates attractive and repulsive forces between nodes to obtain a low-dimensional embedding that maintains the graph topology, node proximities, and graph connectivity information. We demonstrate the effectiveness of our technique on tasks like node classification, link prediction, and visualization on a diverse set of graph network datasets. Our force-directed embedding method outperforms existing unsupervised techniques, highlighting the utility of physics-based simulations for graph representation learning.

\section{Previous Works}

A variety of unsupervised graph embedding techniques have been developed that aim to preserve graph topology in the embedded space without relying on node attributes or labels. We briefly review some representative methods from the major categories discussed in the following.

Matrix factorization-based techniques like Locally Linear Embedding (LLE) \cite{roweis2000nonlinear} and Laplacian Eigenmaps \cite{belkin2003laplacian} aim to factorize a matrix representing node proximity, like the adjacency matrix or Laplacian matrix, to obtain the embeddings. For example, Laplacian Eigenmaps minimize a cost function that penalizes large distances between connected nodes in the embedding. While these methods encode global graph structure, the eigendecomposition becomes expensive for large graphs.

Edge reconstruction techniques like LINE \cite{tang2015line} and SDNE \cite{wang2016structural} directly optimize an objective like edge reconstruction probability or representation likelihood over the embedding vectors. This is efficient but relies solely on direct first-order connections between nodes.

DeepWalk \cite{perozzi2014DeepWalk} and Node2vec \cite{grover2016Node2vec} are random walk-based methods. In random walk-based methods, a set of nodes is sampled from the graph through randomized walks. The random traversal is supposed to reflect the connectivity features of the graph. The embedding of the sampled nodes is then optimized with respect to the co-occurrence probability of neighboring nodes in these walks. Node2vec expands on DeepWalk and uses a different random walk strategy by having a bias for BFS traversal over DFS. Both these works use the skip-gram model of word2vec \cite{mikolov2013distributed} for their optimization objective.

Force-directed algorithms are among the most popular methods for visualizing graph topology and symmetries through 2-dimensional or 3-dimensional layouts. The key idea is to model the graph as a physical system with forces between nodes that determine their layout. Distance-based method \cite{kamada1989algorithm} defines an ideal distance between nodes based on graph distance to minimize the difference between the ideal and actual distances. Electrical force models \cite{fruchterman1991graph} use electrical repulsion forces between nodes. These basic force-directed methods work well for small graphs but often get trapped in local optima for larger graphs. Multilevel approaches \cite{walshaw2001multilevel, gajer2000multi} use previous approaches while coarsening the graph recursively and refining the layout from coarse to fine. This approach helps avoid local minima. N-body simulations \cite{hu2005efficient} use techniques like Barnes-Hut to approximate long-range forces efficiently which allows scaling. In \cite{gajer2000grip} the graph is embedded in a high-dimensional space and projected to 2D or 3D.

While extensively studied for visualization, force-directed techniques have rarely been applied to graph embedding. Rahman et al. in a recent work \cite{rahman2020force2vec,rahman2022scalable} proposed Force2Vec, which uses the spring electrical force model as a loss function. They calculate the repulsive force between all pairs of nodes and the attractive force between only pairs of nodes that are connected by an edge in the graph. In addition, they use stochastic gradient descent with negative sampling for optimization. and optimizations like batching and vector operations to scale training.

Our force-directed embedding method is based on the famous kinetic formula $x=1/2at^2+v_0t+x_0$ to calculate the gradient for embedding at each step. The acceleration is calculated using Newton’s $2^{nd}$ law $a=F/m$ by calculating the sum of all forces induced on each node. We demonstrate through experiments that this holistic approach outperforms existing techniques on several graph analysis tasks.

\section{The Proposed Method}
\subsection{Background}
Let $G(V,E)$ denote a graph with $V$ and $E$ sets representing its sets of $n$ vertices and $m$ edges respectively. 
The embedding process learns the mapping function $f\colon V \longrightarrow\mathbb{R}^d$, $d \ll n$ which maps each element of the set $V$ to a d-dimensional vector space. In other words $f(u_i)=z_i$ for $u_i\in V, z_i \in \mathbb{R}^d$.
A path from node $u_i$ to $u_j$, $i\ne j$ in a graph is a sequence of connected nodes with $u_i$ at the head of the sequence and $u_j$ at the end. A hops-distance matrix $\mathbf{H}\in \mathbb{Z}^{n \times n}_{\geq 0}$ represents the shortest path distance length between nodes. $\mathbf{H}_{ij} \in \mathbb{Z}_{\geq 0}$ represents the length of the shortest path from node $u_i$ to $u_j$ such that each traversal from one node to the next in the sequence counts as one hop. $\mathbf{H}_{ij}=0$ if $i=j$, i.e. link to self counts as zero-hop. $\mathbf{H}_{ij}=\infty$ if there is no path between the two nodes.

\subsection{Our proposed force-directed methodology}
The main premise of our methodology is to embed nodes in the d-dimensional such that the Euclidean distance of node pairs is proportional to their hops-distance and path count between them. We want the nodes that have shorter hop distances in the graph to be embedded proportionally closer in the d-dimensional space. In addition, we want the embedding to capture and reflect other graph structural and connectivity information such as community, bridge, path count, etc.

The force-directed approach comes into frame when considering the kinetic equation model and Newton's second law as follows.

\begin{align}
    x &= \frac{1}{2}at^2+v_0t+x_0 \label{eq:kinetic}\\
    \frac{dx}{dt} &= at+v_0     \label{eq:kinetic_derivative}  \\
    \frac{dx}{dt} &= at_{step}=a \label{eq:kinetic_gradient}\\
    a &= \frac{F}{m} \label{eq:newton2nd}
\end{align}

The equation \eqref{eq:kinetic} shows location $x\in\mathbb{R}^d$ of a mobile object in a d-dimensional vector space at time $t$ with acceleration rate $a\in\mathbb{R}^d$ and initial velocity $v_0\in\mathbb{R}^d$ and initial location $x_0\in\mathbb{R}^d$. In equation \eqref{eq:kinetic_derivative} $\frac{dx}{dt}$ is the rate of change of $x$ at time $t$ and is proportional to acceleration factor $a$.

We wish to use $dx$ from equation \eqref{eq:kinetic_derivative} as the gradient for embedding in a vector space in a step-wise fashion. At the beginning of each step, all nodes are halted which makes $v_0=0$. In addition, we assume timesteps $t_\text{step}=1$ so that we retain an approximately steady acceleration rate during the step and do not render the kinetic equation irrelevant. 

According to Newton's second law \eqref{eq:newton2nd}, the acceleration of an object is directly proportional to the net force acting on the object, and inversely proportional to the mass of the object.

In our proposed method, we assume that all nodes of a graph are mutually exerting attractive and repulsive forces on each other. Therefore, net force imposed on a node can be calculated using \eqref{eq:netforce}. In this equation, $F(u)$ is the net force on node $u$ while $F_\text{attr}(u,v)$ and $F_\text{repl}(u,v)$ are mutual attractive and repulsive forces between nodes $u$ and $v$. 

\begin{align} 
    F(u)&=F_\text{attr}(u)+F_\text{repl}(u) \label{eq:netforce}
\end{align}

We wish larger Euclidean distance between two embeddings to reduce the repulsive force and increase the attraction while larger distances have the opposite effect. In addition, we wish to have similar levels of attraction between node pairs of hops-distance groups. We use \eqref{eq:forceattr} and \eqref{eq:forcerepl} equations for attractive and repulsive force models respectively. In these equations $\lVert z_{uv} \rVert = \lVert z_v-z_u \rVert$ is the Euclidean distance between the embeddings of two nodes $u$ and $v$, $h_{uv}$ is their hops-distance and $0<\alpha<1$. $V$ is the set of all nodes while $N_u^{(h)}$ is the set of neighbors of $u$ at $h$-hops distance. $unit_{uv} = \frac{z_{uv}}{\lVert z_{uv} \rVert}$ is the unit vector along which the tension between $u$ and $v$ produces a force.

\begin{align}
    F_\text{attr}(u) &=  \sum_{h=1}^{\infty} {\frac{1}{|N_u^{(h)}| } \sum_{v \in N_u^{(h)}}  {\alpha^{h_{uv}-1} \lVert z_{uv} \rVert}} \times unit_{uv} \label{eq:forceattr} \\
    F_\text{repl}(u) &= \frac{1}{|V| } \sum_{v \in V} {h_{uv} e^{-\lVert z_{uv} \rVert}}  \times unit_{uv} \label{eq:forcerepl}
\end{align}

All in all, by replacing \eqref{eq:netforce} the gradient equation derived in \eqref{eq:kinetic_gradient} we arrive at equation \eqref{eq:gradient}. In this equation, $dz_u$ is the gradient of embedding of node $u$. For mass of node $u$ we let $m_u=\deg u$. The intuition behind this is that the embedding of a node with more edges should be discounted on its gradient, increasing its inertia to alteration.

\begin{align} 
    dz_u&= \frac{F(u)}{m_u}
    \label{eq:gradient}
\end{align}

\subsection{Avoiding local optima}
A force-directed system may converge to a local optimum point. To circumvent the local optima we adopt a random drop strategy. For each calculated gradient, we randomly pick some dimensions with 0.5 probability and zero them out. This strategy enhances the embedding results as shown later in this manuscript.

\subsection{The algorithm}
The algorithm~\ref{alg:embedding} shows the procedure of our force-directed embedding. The algorithm starts by initializing random embedding vectors for each node. It then starts a loop where it calculates the net force imposed on each node to obtain its embedding gradient. Finally, it applies a random drop function on the gradients and updates the corresponding embedding vectors. The loop stops when the force-directed system enters an equilibrium state where the sum of the magnitude of net forces on all nodes reaches a minimum level. The two inner loops can be parallelized using vector operations.

\begin{algorithm}
\caption{Force-directed Graph Embedding}
\label{alg:embedding}
\begin{algorithmic}[1]
\State Randomly initialize all $z_u$, $u \in V$
\While{$\sum_{u\in V}\lVert F(u) \rVert >\epsilon$}
    \ForAll{$u\in V$} \Comment{Calculate the gradients first}
    \State $dz_u=\frac{F(u)}{\deg u}$
    \EndFor
    \ForAll{$u\in V$} \Comment{Update embeddings}
    \State $z_u+=$\textbf{RandomDrop}($dz_u$)
    \EndFor
\EndWhile
\end{algorithmic}
\end{algorithm}

\subsection{Algorithm complexity}
In this algorithm, forces between each pair of nodes are to be calculated. Therefore, the time complexity of this algorithm is $O(n^2)$. But since this algorithm is parallelizeable, it is possible to do a tradeoff between time and space complextity by using batching techniques.

\section{Evaluation Results}
In this section, we discuss the evaluation results of our force-directed methodology and compare it against the best available methods prior to this work. All the evaluations and comparisons are performed on 128-dimensional graph embeddings generated by the target methods. Two major graph tasks for evaluating the quality of a graph embedding are node classification and link predictions. They are thoroughly discussed in their corresponding sections. In these evaluations we used \texttt{RandomForestClassifier}, the random forest implementation from Scikit-learn package\cite{sklearn_api, scikit-learn} version 1.2.2. Along with the default parameters of \texttt{RandomForestClassifier}, we used 2\% as the minimum fraction of samples for tree node splitting. The rationale behind using random forest was its being one of the best-performing algorithms in classification.

The implementation of this work and the evaluation methodology discussed in this section can be accessed at \url{https://github.com/HessamLa/forcedirected}.

\subsection{Datasets and baseline methods}
The following graph datasets were used for evaluating various embedding methods. 
\begin{itemize}
\item Cora \cite{sen2008collective}; This dataset consists of 2708 scientific publications, each classified into one of seven classes. Every pair of citing and cited publications in this dataset is a link which accounts for 5429 citations.
\item CiteSeer \cite{sen2008collective}; This dataset contains 3312 scientific publications, each classified into one of six topics. The publications have 4732 citations between them, forming a citation network.
\item Pub-Med \cite{namata2012query}; The Pubmed Diabetes dataset contains 19717 scientific publications related to diabetes, sourced from the PubMed database. These publications are categorized into one of three classes. Its citation network consists of 44338 citation links.
\item Ego-Facebook \cite{leskovec2012learning}; This is a Facebook friends network collected from a survey participants, representing ego-networks centered around individual users. 10 users provided access to their friend connections on Facebook, and manually labeled groups of friends into "circles" or communities. In total, the Ego-Facebook dataset comprises 4039 nodes representing friends, connected by 88234 links, and organized into 193 ground-truth circles labeled by the ego users. On average, each ego network has around 400 friends, divided into 19 circles of 22 friends each. The circles capture social contexts such as university friends, relatives, sports teams, etc. By capturing full ego-networks with ground truth community assignments, the Ego-Facebook dataset enables analysis of community detection and graph mining algorithms, providing a benchmark for methods aiming to identify overlapping and hierarchical circles in social networks centered around individuals.

\item Wiki\footnote{\url{https://github.com/thunlp/MMDW/} accessed on July 28.2023}; This data is generated from scraping Wikipedia pages. It consists of 2405 Wikipedia pages and 17981 hyperlink redirections between them. The pages are exclusively classified into 19 categories.
\item CORA-Full \cite{bojchevski2017deep}; This dataset (named CORA in the original paper) is a citation network containing 19793 scientific publications classified into 70 different classes. Each publication is represented as a 0/1-valued word vector indicating the absence/presence of 1433 unique dictionary words extracted from the paper abstracts. In total, the dataset contains 65311 links representing citations between the publications. The rich feature representation coupled with the citation linkage structure makes CORA well-suited for analyzing representation learning techniques on node classification tasks.
\end{itemize}

All of the above graphs were used for the link prediction task. For node classification however, node labels are required. Therefore, the Ego-Facebook graph did not contain node labeling and was not used for node classification.

We compare our embedding method against LINE, DeepWalk, Node2vec and GraphSAGE methods in terms of accuracy and macro F-1 scores of link prediction and node classification tasks.

\input{figs-accuracy}

\input{tables}

\subsection{Link Prediction}
The objective of link prediction is to predict if a link exists between two given nodes. Here, embeddings of two nodes are passed to the predictor and a boolean output is obtained which indicates exsitence or non-existence of a link between the two input embeddings. 

For the link prediction task in this evaluation, the negative sampling technique is used. 80\% of the available node pairs are used for training and 20\% for testing. For each (method, dataset) pair the evaluation is performed 7 times and the average is reported here. Tables \ref{tab:linkpred-acc} and \ref{tab:linkpred-f1score} show the accuracy and macro F-1 score of the link prediction task with random forest. Our Force-Directed method outperformed the best available unsupervised embedding methods while maintaining a low standard deviation for the 7 evaluation runs.

\subsection{Node classification task}
The objective of node classification is to classify a node label. For this task, the classifier was trained on 80\% of the nodes and tested on the remaining 20\%. We did 7 runs of evaluation for this task and reported the average in tables \ref{tab:nodeclass-acc} and \ref{tab:nodeclass-f1score} for accuracy and macro F-1 scores respectively. As shown here Our method outperformed others on all datasets.

\subsection{Embedding progression}
The Force-Directed embedding starts from randomly generated embeddings in the vector space. As the algorithm progresses, the Euclidean distance between node pairs embeddings for each group of h-hops distant nodes gradually converge to a point where on average, they reflect hops distance between pairs with negligible standard deviation. 

\begin{figure*}[ht]
\centering
    \includegraphics[width=0.99\textwidth]{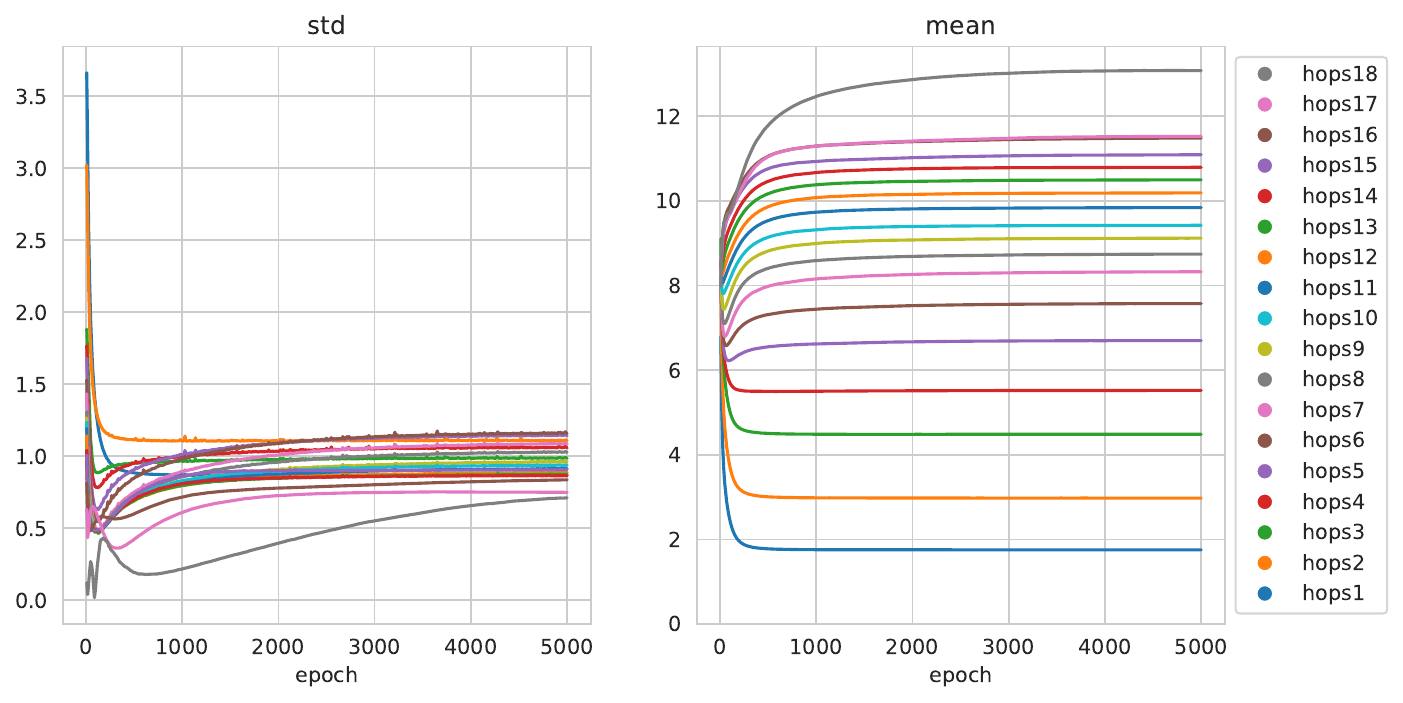}
    \caption{Progression of mean and standard deviation of Euclidean distances for Pub-Med graph. The set of all node pairs is partitioned based on h-hops distance. Then the mean and standard deviation of Euclidean distance of node pairs embeddings for each partition is calculated. For example, "hops1" is the partition of node pairs that are immediate neighbors, while "hops5" is the partition of node pairs that are 5 hops away on the graph.}
\label{fig:pairdist}
\end{figure*}

Figure \ref{fig:pairdist} shows the progression of Euclidean distances when using the Force-Directed method for generating embeddings for the Pub-Med graph. In this figure, "hops(n)" indicates the set of node pairs at n-hop distance. For example, "hops1" is the set of node pairs that are immediate neighbors, while "hops7" indicates the set of pairs that are 7 hops away. The maximum distance on this graph is 18 hops. This figure shows that nodes in closer vicinity, i.e. in the same cluster, quickly converge to stable distances. However, Euclidean distances of further node pairs take more iterations to converge, indicating that clusters take more time to converge to stable distances. This observation is pointed out in the Qualitative analysis subsection.

\subsection{Qualitative analysis}
Figure \ref{fig:embeddings-plot} shows the 2D representation of Ego-Facebook network embeddings generated by the three best-performing methods, Force-Directed, DeepWalk, and Node2vec. For producing a 2D visualization of the embeddings, the 128-dimensional embeddings are reduced to 2-dimensional using PCA. In these graphs, nodes with higher degrees have brighter colors and are depicted larger. As seen in these graphs, while all methods are performing rather well in keeping clusters of interconnected nodes together, Force-Directed embedding output does an even better job of producing a more pronounced distance between clusters by distancing less interconnected clusters. In addition, node types in terms of their structural role such as hub, connector, and peripheral nodes. Hub nodes have a high degree and are connected to many other nodes in the network. They act as highly connected centers for information spread. Connector nodes link hubs from different communities, helping spread information between clusters. Peripheral nodes are at the edges of a cluster with fewer connections than inner members.

\setcounter{page}{5}

\input{figs-embeddings-plot}

\section{Conclusion}
We presented a novel graph embedding technique using a force-directed approach that simulates physics-based forces between nodes to obtain embeddings preserving graph topology and connectivity patterns. Our intuitive method demonstrated strong performance on tasks like node classification and link prediction compared to existing unsupervised techniques. Important future work includes optimizing the complexity of the algorithm and scaling the embedding method to massive graphs with billions of edges. A niche on researching optimal force functions with Force-Directed embedding method is foreseeable.

\bibliographystyle{unsrt} 
\bibliography{refs-gnn} 


\end{document}

%% file: figs-accuracy.tex
\begin{figure}[t!]
\centering
\includegraphics[width=0.45\textwidth]{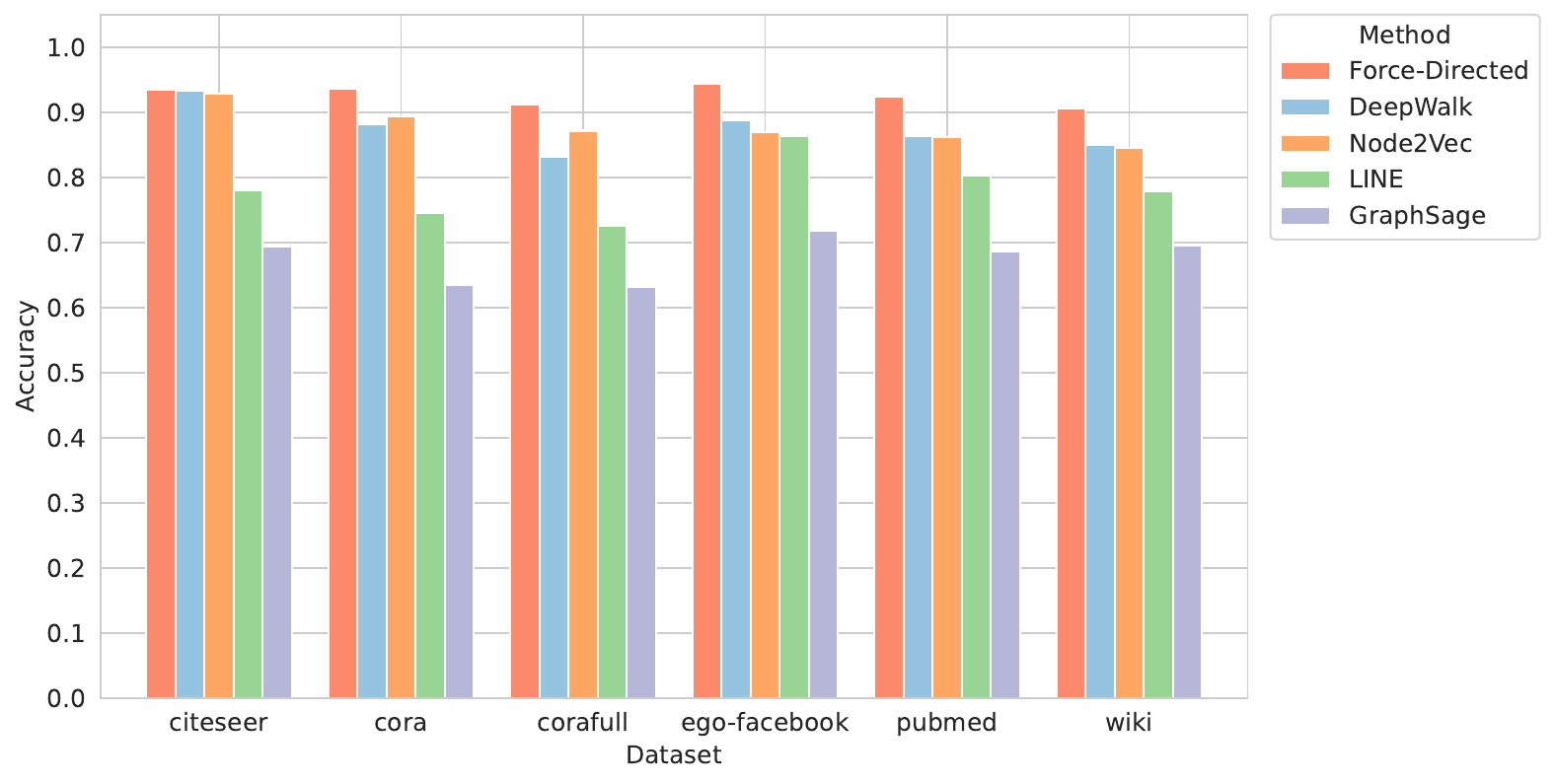}
\caption{Accuracy comparison of link prediction task with decision tree classifier.}
\label{fig:linkpred-acc-rf}

\includegraphics[width=0.45\textwidth]{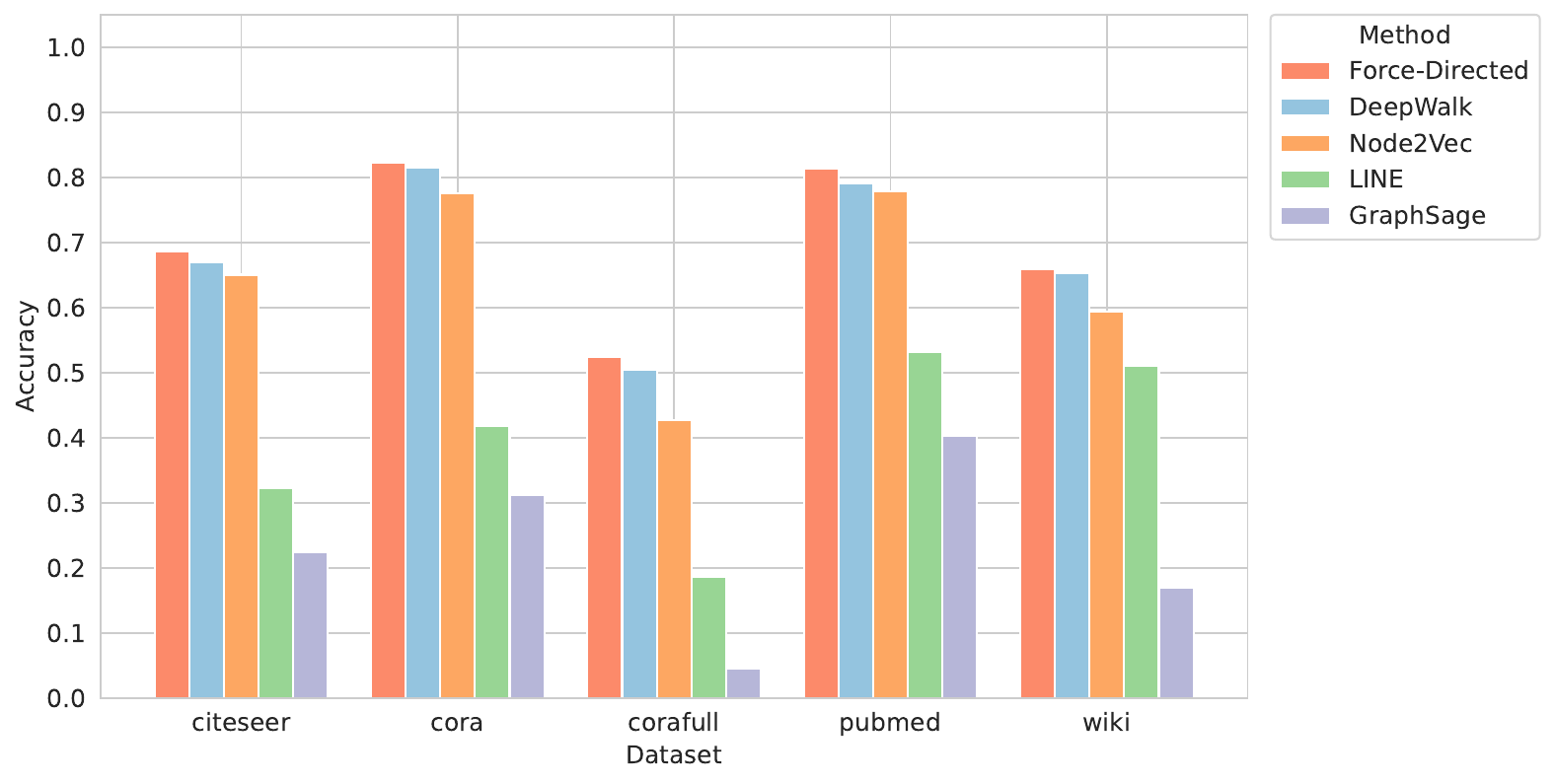}

\caption{Accuracy comparison of node classification task with random forest classifier.}
\label{fig:nodeclass-acc-rf}
\end{figure}

%% file: tables.tex
\begin{table*}[t]
\caption{Accuracy percentage of graph embedding algorithms in link prediction task using random forest classifier}
\label{tab:linkpred-acc}
\begin{center}
\begin{tabular}{|c|c c c c c c|}
\hline
\textbf{Embedding}&\multicolumn{6}{|c|}{\textbf{Dataset}} \\
\cline{2-7} 
\textbf{Method} & \textbf{citeseer} & \textbf{cora} & \textbf{corafull} & \textbf{ego-facebook} & \textbf{pubmed} & \textbf{wiki} \\
\hline
Force-Directed & \textbf{93.51}$\pm$0.38\% & \textbf{93.62}$\pm$0.57\% & \textbf{91.15}$\pm$0.30\% & \textbf{94.32}$\pm$0.10\% & \textbf{92.47}$\pm$0.30\% & \textbf{90.56}$\pm$0.26\% \\
DeepWalk & 93.35$\pm$0.86\% & 88.15$\pm$0.56\% & 83.22$\pm$0.45\% & 88.86$\pm$0.10\% & 86.41$\pm$0.46\% & 84.97$\pm$0.50\% \\
Node2Vec & 92.92$\pm$0.53\% & 89.42$\pm$0.60\% & 87.20$\pm$0.29\% & 87.03$\pm$0.11\% & 86.18$\pm$0.24\% & 84.49$\pm$0.29\% \\
GraphSage & 69.36$\pm$1.04\% & 63.44$\pm$0.77\% & 63.16$\pm$0.18\% & 71.80$\pm$0.26\% & 68.67$\pm$0.20\% & 69.55$\pm$0.70\% \\
LINE & 78.00$\pm$0.64\% & 74.57$\pm$1.19\% & 72.52$\pm$0.33\% & 86.33$\pm$0.13\% & 80.35$\pm$0.32\% & 77.94$\pm$0.57\% \\
\hline
\end{tabular}
\end{center}
\end{table*}

\begin{table*}[t]
\caption{Macro F1-score of graph embedding algorithms in link prediction task using random forest classifier}
\label{tab:linkpred-f1score}
\begin{center}
\begin{tabular}{|c|c c c c c c|}
\hline
\textbf{Embedding}&\multicolumn{6}{|c|}{\textbf{Dataset}} \\
\cline{2-7} 
\textbf{Method} & \textbf{citeseer} & \textbf{cora} & \textbf{corafull} & \textbf{ego-facebook} & \textbf{pubmed} & \textbf{wiki} \\
\hline
Force-Directed & \textbf{0.93}$\pm$0.003 & \textbf{0.93}$\pm$0.005 & \textbf{0.91}$\pm$0.003 & \textbf{0.94}$\pm$0.001 & \textbf{0.92}$\pm$0.003 & \textbf{0.90}$\pm$0.002 \\
DeepWalk & 0.93$\pm$0.008 & 0.88$\pm$0.005 & 0.83$\pm$0.004 & 0.88$\pm$0.001 & 0.86$\pm$0.004 & 0.84$\pm$0.005 \\
Node2Vec & 0.92$\pm$0.005 & 0.89$\pm$0.006 & 0.87$\pm$0.003 & 0.86$\pm$0.001 & 0.86$\pm$0.002 & 0.84$\pm$0.002 \\
GraphSage & 0.69$\pm$0.010 & 0.63$\pm$0.007 & 0.63$\pm$0.001 & 0.71$\pm$0.002 & 0.68$\pm$0.002 & 0.68$\pm$0.006 \\
LINE & 0.76$\pm$0.007 & 0.73$\pm$0.013 & 0.71$\pm$0.003 & 0.86$\pm$0.001 & 0.80$\pm$0.003 & 0.77$\pm$0.005 \\
\hline
\end{tabular}
\end{center}
\end{table*}

\begin{table*}[t]
\caption{Accuracy percentage of graph embedding algorithms in node classification task using random forest classifier}
\label{tab:nodeclass-acc}
\begin{center}
\begin{tabular}{|c|c c c c c|}
\hline
\textbf{Embedding}&\multicolumn{5}{|c|}{\textbf{Dataset}} \\
\cline{2-6} 
\textbf{method} & \textbf{citeseer} & \textbf{cora} & \textbf{corafull} & \textbf{pubmed} & \textbf{wiki} \\
\hline
Force-Directed & \textbf{68.64}$\pm$1.49\% & \textbf{82.34}$\pm$1.38\% & \textbf{52.41}$\pm$1.02\% & \textbf{81.30}$\pm$0.68\% & \textbf{65.87}$\pm$1.41\% \\
DeepWalk  & 66.92$\pm$1.57\% & 81.57$\pm$1.65\% & 50.43$\pm$0.62\% & 79.10$\pm$0.61\% & 65.34$\pm$1.71\% \\
Node2Vec  & 65.07$\pm$2.36\% & 77.62$\pm$2.16\% & 42.72$\pm$0.64\% & 77.93$\pm$0.53\% & 59.42$\pm$2.08\% \\
GraphSAGE & 22.37$\pm$1.18\% & 31.28$\pm$1.28\% &  4.58$\pm$0.34\% & 40.26$\pm$0.32\% & 16.98$\pm$1.51\% \\
LINE      & 32.28$\pm$1.30\% & 41.77$\pm$1.60\% & 18.61$\pm$0.43\% & 53.20$\pm$0.81\% & 51.11$\pm$2.10\% \\
\hline
\end{tabular}
\end{center}
\end{table*}

\begin{table*}[t]
\caption{Macro F1-score of graph embedding algorithms in node classification task using random forest classifier}
\label{tab:nodeclass-f1score}
\begin{center}
\begin{tabular}{|c|c c c c c|}
\hline
\textbf{Embedding}&\multicolumn{5}{|c|}{\textbf{Dataset}} \\
\cline{2-6}
\textbf{method} & \textbf{citeseer} & \textbf{cora} & \textbf{corafull} & \textbf{pubmed} & \textbf{wiki} \\
\hline
Force-Directed & \textbf{0.62}$\pm$0.011 & \textbf{0.81}$\pm$0.018 & \textbf{0.34}$\pm$0.013 & \textbf{0.79}$\pm$0.008 & \textbf{0.50}$\pm$0.026 \\
DeepWalk & 0.59$\pm$0.012 & 0.80$\pm$0.018 & \textbf{0.34}$\pm$0.006 & 0.76$\pm$0.006 & 0.46$\pm$0.018 \\
Node2Vec & 0.57$\pm$0.022 & 0.76$\pm$0.027 & 0.24$\pm$0.006 & 0.75$\pm$0.004 & 0.39$\pm$0.016 \\
GraphSage & 0.14$\pm$0.011 & 0.06$\pm$0.002 & 0.00$\pm$0.001 & 0.29$\pm$0.002 & 0.03$\pm$0.004 \\
LINE & 0.25$\pm$0.012 & 0.26$\pm$0.016 & 0.08$\pm$0.003 & 0.39$\pm$0.004 & 0.35$\pm$0.010 \\
\hline
\end{tabular}
\end{center}
\end{table*}

%% file: figs-embeddings-plot.tex
\begin{figure*}[t]
\centering
\minipage{0.30\textwidth}
    \includegraphics[width=\linewidth]{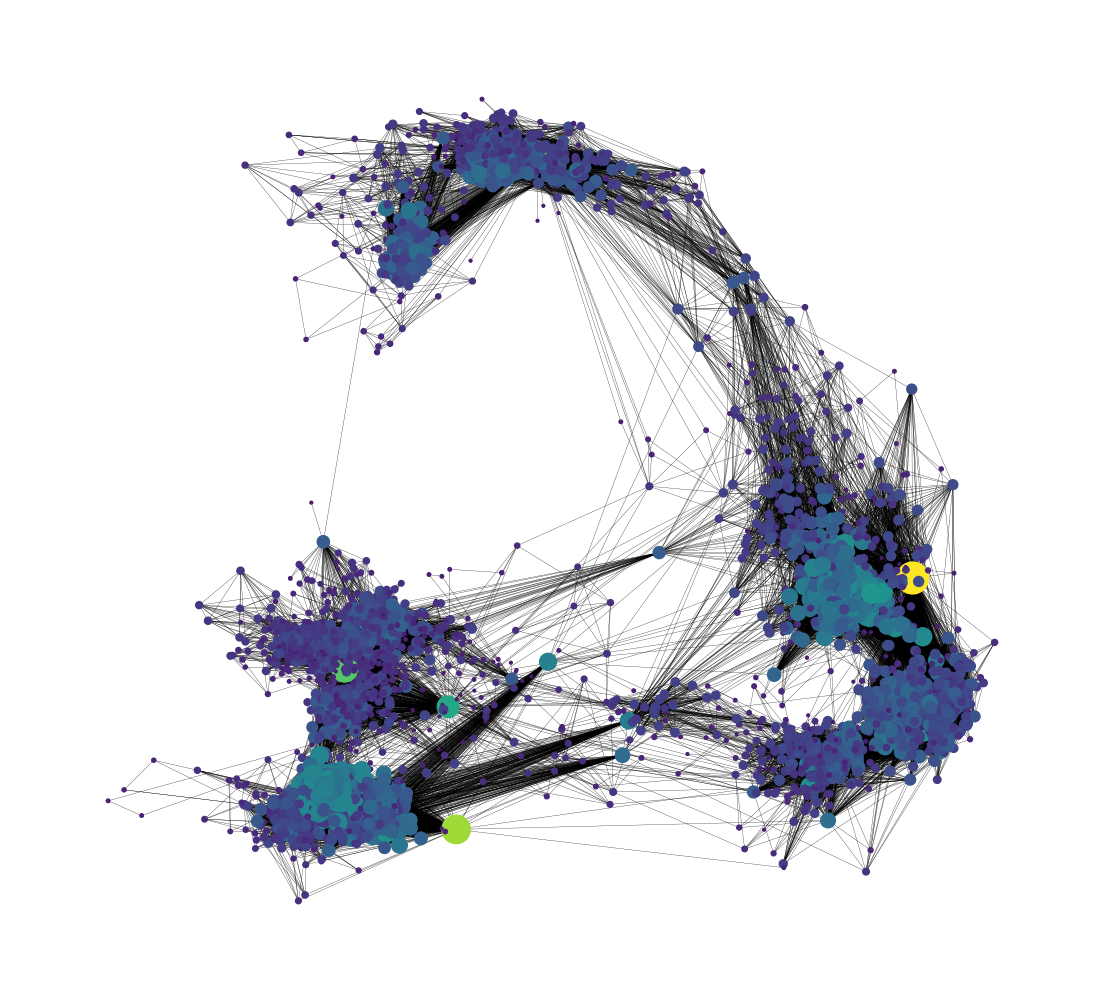}
\endminipage\hfill
\minipage{0.36\textwidth}
    \includegraphics[width=\linewidth]{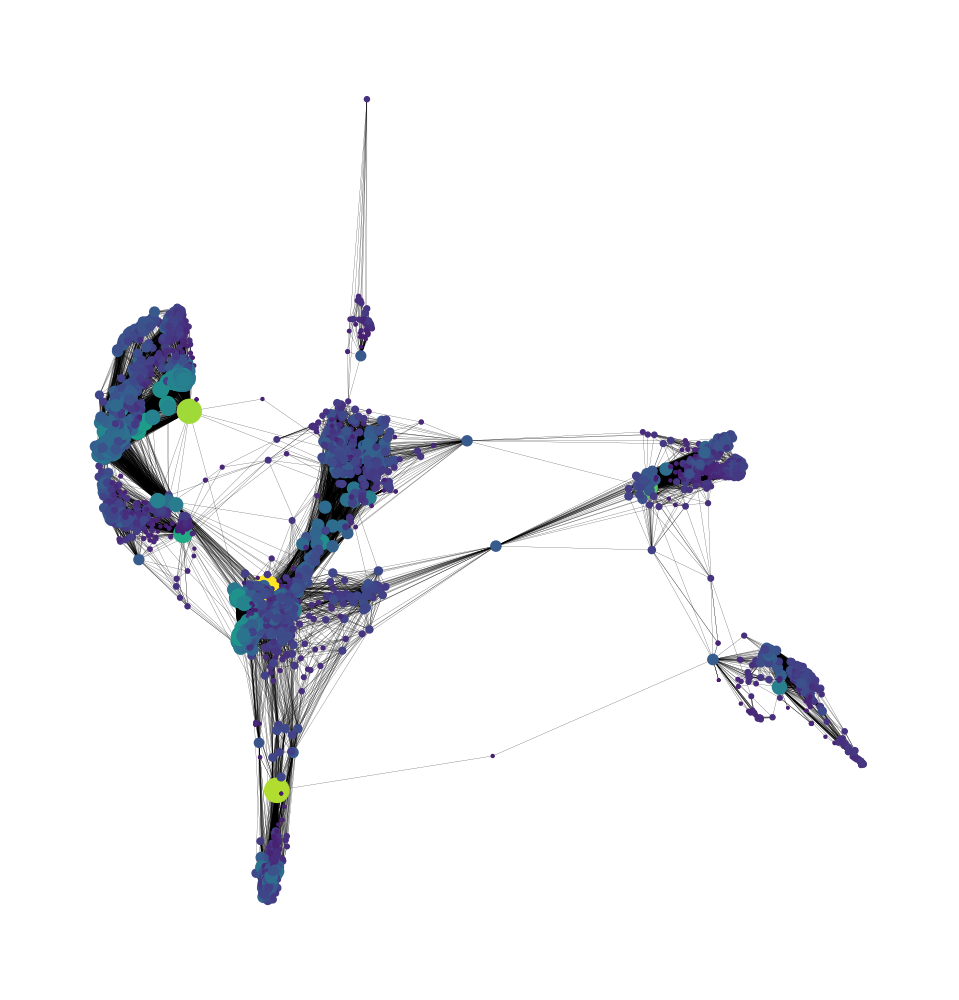}
\endminipage\hfill
\minipage{0.30\textwidth}
    \includegraphics[width=\linewidth]{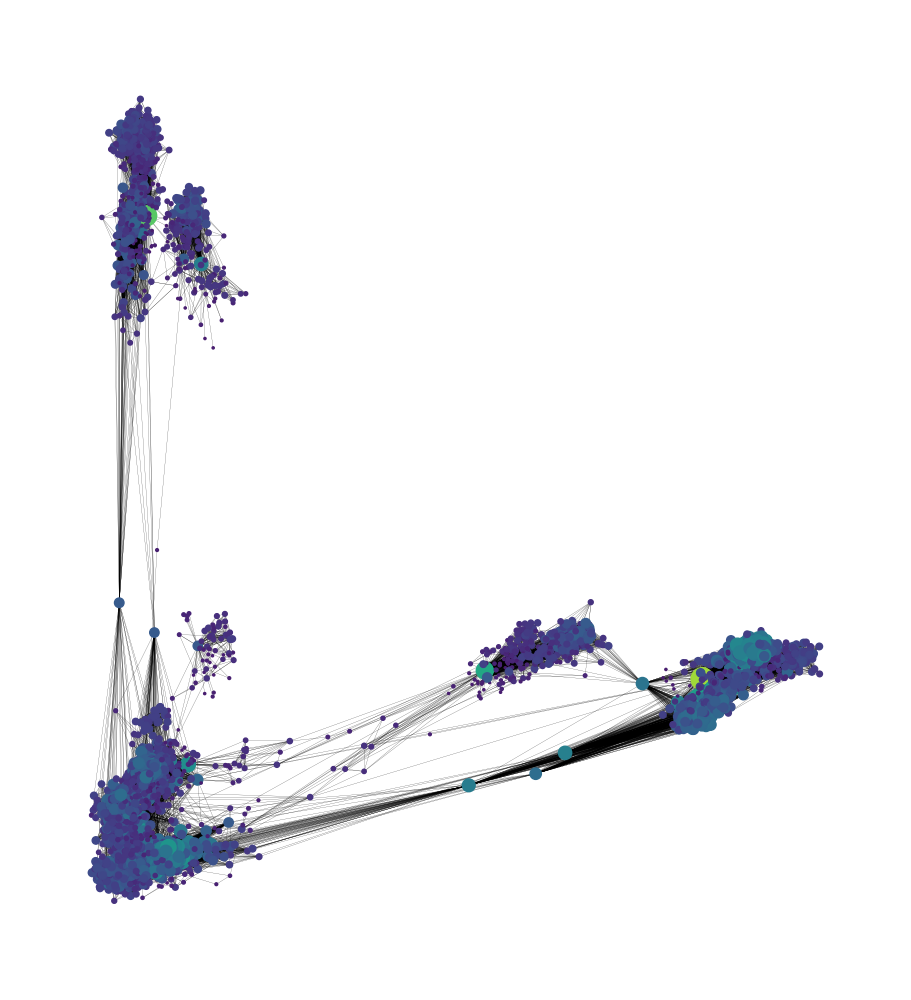}
\endminipage\hfill
\caption{Visual comparison of embedding of Ego-Facebook graph network using Node2vec (left), Force-Directed (center) and DeepWalk (right) methods. Brighter color and larger size of a node indicates its larger degree.}
\label{fig:embeddings-plot}
\end{figure*}

%% file: main.bbl
\begin{thebibliography}{10}

\bibitem{perozzi2014DeepWalk}
Bryan Perozzi, Rami Al-Rfou, and Steven Skiena.
\newblock Deepwalk: Online learning of social representations.
\newblock In {\em Proceedings of the 20th ACM SIGKDD international conference
  on Knowledge discovery and data mining}, pages 701--710, 2014.

\bibitem{cao2015grarep}
Shaosheng Cao, Wei Lu, and Qiongkai Xu.
\newblock Grarep: Learning graph representations with global structural
  information.
\newblock In {\em Proceedings of the 24th ACM international on conference on
  information and knowledge management}, pages 891--900, 2015.

\bibitem{grover2016Node2vec}
Aditya Grover and Jure Leskovec.
\newblock node2vec: Scalable feature learning for networks.
\newblock In {\em Proceedings of the 22nd ACM SIGKDD international conference
  on Knowledge discovery and data mining}, pages 855--864, 2016.

\bibitem{tang2015line}
Jian Tang, Meng Qu, Mingzhe Wang, Ming Zhang, Jun Yan, and Qiaozhu Mei.
\newblock Line: Large-scale information network embedding.
\newblock In {\em Proceedings of the 24th international conference on world
  wide web}, pages 1067--1077, 2015.

\bibitem{wang2016structural}
Daixin Wang, Peng Cui, and Wenwu Zhu.
\newblock Structural deep network embedding.
\newblock In {\em Proceedings of the 22nd ACM SIGKDD international conference
  on Knowledge discovery and data mining}, pages 1225--1234, 2016.

\bibitem{hamilton2016diachronic}
William~L Hamilton, Jure Leskovec, and Dan Jurafsky.
\newblock Diachronic word embeddings reveal statistical laws of semantic
  change.
\newblock {\em arXiv preprint arXiv:1605.09096}, 2016.

\bibitem{roweis2000nonlinear}
Sam~T Roweis and Lawrence~K Saul.
\newblock Nonlinear dimensionality reduction by locally linear embedding.
\newblock {\em science}, 290(5500):2323--2326, 2000.

\bibitem{belkin2003laplacian}
Mikhail Belkin and Partha Niyogi.
\newblock Laplacian eigenmaps for dimensionality reduction and data
  representation.
\newblock {\em Neural computation}, 15(6):1373--1396, 2003.

\bibitem{mikolov2013distributed}
Tomas Mikolov, Ilya Sutskever, Kai Chen, Greg~S Corrado, and Jeff Dean.
\newblock Distributed representations of words and phrases and their
  compositionality.
\newblock {\em Advances in neural information processing systems}, 26, 2013.

\bibitem{kamada1989algorithm}
Tomihisa Kamada, Satoru Kawai, et~al.
\newblock An algorithm for drawing general undirected graphs.
\newblock {\em Information processing letters}, 31(1):7--15, 1989.

\bibitem{fruchterman1991graph}
Thomas~MJ Fruchterman and Edward~M Reingold.
\newblock Graph drawing by force-directed placement.
\newblock {\em Software: Practice and experience}, 21(11):1129--1164, 1991.

\bibitem{walshaw2001multilevel}
Chris Walshaw.
\newblock A multilevel algorithm for force-directed graph drawing.
\newblock In {\em Graph Drawing: 8th International Symposium, GD 2000 Colonial
  Williamsburg, VA, USA, September 20--23, 2000 Proceedings 8}, pages 171--182.
  Springer, 2001.

\bibitem{gajer2000multi}
Pawel Gajer, Michael~T Goodrich, and Stephen~G Kobourov.
\newblock A multi-dimensional approach to force-directed layouts of large
  graphs.
\newblock In {\em International Symposium on Graph Drawing}, pages 211--221.
  Springer, 2000.

\bibitem{hu2005efficient}
Yifan Hu.
\newblock Efficient, high-quality force-directed graph drawing.
\newblock {\em Mathematica journal}, 10(1):37--71, 2005.

\bibitem{gajer2000grip}
Pawel Gajer and Stephen~G Kobourov.
\newblock Grip: Graph drawing with intelligent placement.
\newblock In {\em International Symposium on Graph Drawing}, pages 222--228.
  Springer, 2000.

\bibitem{rahman2020force2vec}
Md~Khaledur Rahman, Majedul~Haque Sujon, and Ariful Azad.
\newblock Force2vec: Parallel force-directed graph embedding.
\newblock In {\em 2020 IEEE International Conference on Data Mining (ICDM)},
  pages 442--451. IEEE, 2020.

\bibitem{rahman2022scalable}
Md~Khaledur Rahman, Majedul~Haque Sujon, and Ariful Azad.
\newblock Scalable force-directed graph representation learning and
  visualization.
\newblock {\em Knowledge and Information Systems}, 64(1):207--233, 2022.

\bibitem{sklearn_api}
Lars Buitinck, Gilles Louppe, Mathieu Blondel, Fabian Pedregosa, Andreas
  Mueller, Olivier Grisel, Vlad Niculae, Peter Prettenhofer, Alexandre
  Gramfort, Jaques Grobler, Robert Layton, Jake VanderPlas, Arnaud Joly, Brian
  Holt, and Ga{\"{e}}l Varoquaux.
\newblock {API} design for machine learning software: experiences from the
  scikit-learn project.
\newblock In {\em ECML PKDD Workshop: Languages for Data Mining and Machine
  Learning}, pages 108--122, 2013.

\bibitem{scikit-learn}
F.~Pedregosa, G.~Varoquaux, A.~Gramfort, V.~Michel, B.~Thirion, O.~Grisel,
  M.~Blondel, P.~Prettenhofer, R.~Weiss, V.~Dubourg, J.~Vanderplas, A.~Passos,
  D.~Cournapeau, M.~Brucher, M.~Perrot, and E.~Duchesnay.
\newblock Scikit-learn: Machine learning in {P}ython.
\newblock {\em Journal of Machine Learning Research}, 12:2825--2830, 2011.

\bibitem{sen2008collective}
Prithviraj Sen, Galileo Namata, Mustafa Bilgic, Lise Getoor, Brian Galligher,
  and Tina Eliassi-Rad.
\newblock Collective classification in network data.
\newblock {\em AI magazine}, 29(3):93--93, 2008.

\bibitem{namata2012query}
Galileo Namata, Ben London, Lise Getoor, Bert Huang, and U~Edu.
\newblock Query-driven active surveying for collective classification.
\newblock In {\em 10th international workshop on mining and learning with
  graphs}, volume~8, page~1, 2012.

\bibitem{leskovec2012learning}
Jure Leskovec and Julian Mcauley.
\newblock Learning to discover social circles in ego networks.
\newblock {\em Advances in neural information processing systems}, 25, 2012.

\bibitem{bojchevski2017deep}
Aleksandar Bojchevski and Stephan G{\"u}nnemann.
\newblock Deep gaussian embedding of graphs: Unsupervised inductive learning
  via ranking.
\newblock {\em arXiv preprint arXiv:1707.03815}, 2017.

\end{thebibliography}
